\documentclass[11pt,a4paper]{article}
\usepackage[hyperref]{acl2019}
\usepackage{amsmath}
\usepackage{times}
\usepackage{latexsym}
\usepackage{multicol}
\usepackage{multirow}
\usepackage{algorithm}
\usepackage[noend]{algpseudocode}
\usepackage[colorinlistoftodos]{todonotes}
\usepackage{comment}
\usepackage{enumitem}
\usepackage{breqn}
\usepackage{tabularx}
\usepackage{arydshln}
\usepackage{graphicx}
\usepackage{makecell}
\usepackage{caption}
\usepackage{subcaption}
\usepackage{ dsfont }
\usepackage{ textcomp }

\graphicspath{ {Pictures/} }
\setlist{nosep}

\usepackage{array}
\newcolumntype{L}[1]{>{\raggedright\let\newline\\\arraybackslash\hspace{0pt}}m{#1}}
\newcolumntype{C}[1]{>{\centering\let\newline\\\arraybackslash\hspace{0pt}}m{#1}}
\newcolumntype{R}[1]{>{\raggedleft\let\newline\\\arraybackslash\hspace{0pt}}m{#1}}
\newcommand*{\Scale}[2][4]{\scalebox{#1}{$#2$}}%

\usepackage{url}

\author{Daphne Ippolito$^{\star}$\quad Reno Kriz$^{\star}$\quad Maria Kustikova\quad Jo\~{a}o Sedoc\quad Chris Callison-Burch\\
  $^{\star}$Authors contributed equally\\
  University of Pennsylvania\\
  {\tt \{daphnei,rekriz,mkust,joao,ccb\}@seas.upenn.edu}
}

\aclfinalcopy % Uncomment this line for the final submission
\newcommand{\td}[1]{#1\textdagger}

\title{Comparison of Diverse Decoding Methods from Conditional Language Models}

\date{}
\begin{document}
\maketitle

\begin{abstract}
    While conditional language models have greatly improved in their ability to output high-quality natural language, many NLP applications benefit from being able to generate a diverse set of candidate sequences.
    Diverse decoding strategies aim to, within a given-sized candidate list, cover as much of the space of high-quality outputs as possible, leading to improvements for tasks that re-rank and combine candidate outputs.
    Standard decoding methods, such as beam search, optimize for generating high likelihood sequences rather than diverse ones, though recent work has focused on increasing diversity in these methods.
    In this work, we perform an extensive survey of decoding-time strategies for generating diverse outputs from conditional language models.
    We also show how diversity can be improved without sacrificing quality by over-sampling additional candidates, then filtering to the desired number.
    \end{abstract}

\section{Introduction}
Conditional neural language models, which train a neural net to map from one sequence to another, have had enormous success in natural language processing tasks such as machine translation \cite{sutskever2014sequence,Luong2015EffectiveAT}, text summarization \cite{Nallapati2016AbstractiveTS}, and dialog systems \cite{Vinyals2015ANC}.
These models output a probability distribution over the next token in the output sequence given the input and the previously predicted tokens.
Since computing the overall most likely output sequence is intractable, early work in neural machine translation found that beam search is an effective strategy to heuristically sample sufficiently likely sequences from these probabilistic models \cite{sutskever2014sequence}.
However, for more open-ended tasks, beam search is ill-suited to generating a set of diverse candidate sequences; this is because candidates outputted from a large-scale beam search often only differ by punctuation and minor morphological variations \cite{li2016mutual}.

The term ``diversity" has been defined in a variety of ways in the literature, with some using it as a synonym for sentence interestingness or unlikeliness \citep{tatsunori2019unifying}, and others considering it a measure of how different two or more sentences are from each other \citep{vijayakumar2016diverse,gimpel2013systematic}. We take the latter approach, and define diversity as the ability of a generative method to create a set of possible outputs that are each valid given the input, but vary as widely as possible in terms of word choice, topic, and meaning.

\begin{figure}
    \scriptsize
    \begin{subfigure}[l]{3.3cm}
    \includegraphics[width=3.3cm]{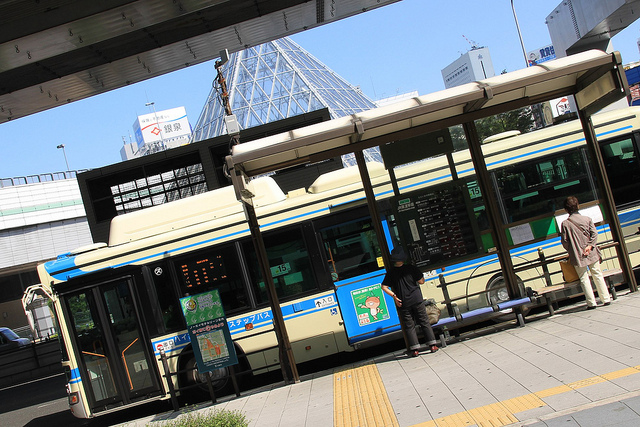}
    \end{subfigure}
    \begin{subfigure}[l]{3cm}
    \begin{tabular}{l}
\textbf{Beam Search}  \\
A bus is stopped at a bus stop. \\
A bus is parked at a bus stop. \\
A bus stopped at a bus stop in a city. \\
A bus stopped at a bus stop at a bus stop. \\
A bus that is parked in front of a building. \\
\hline
\textbf{Random Sampling}  \\
A bus parked at a bus stop at a bus stop.  \\
There is a bus that is at the station. \\
A man standing by a bus in a city.  \\
A bus pulling away from the train station. \\
A bus stopped at a stop on the sunny day. \\
    \end{tabular}
    \end{subfigure}
    \caption{An image with the top five captions from standard beam search and from random sampling. Note the latter set is more diverse but lower quality.}
    \label{first_example}
\end{figure}

There are a number of reasons why it is desirable to produce a set of diverse candidate outputs for a given input.
For example, in collaborative story generation, the system makes suggestions to a user for what they should write next \cite{clark2018creative}.
In these settings, it would be beneficial to show the user multiple different ways to continue their story.
In image captioning, any one sentence-long caption is probably missing some information about the image.
\citet{krause2017hierarchical} show how a set of diverse sentence-length image captions can be transformed into an entire paragraph about the image.
Lastly, in applications that involve reranking candidate sequences, the reranking algorithms are more effective when the input sequences are diverse.
Reranking diverse candidates has been shown to improve results in both open dialog and machine translation \cite{li2016diversity,li2016mutual,gimpel2013systematic}. 
Furthermore, in open-ended dialog, the use of reranking to personalize a model's responses for each user is a promising research direction \cite{Choudhary2017DomainAN}.

With these sorts of applications in mind, a variety of alternatives and extensions to beam search have been proposed which seek to produce a set of diverse candidate responses instead of a single high likelihood one \cite{li2016diversity,vijayakumar2016diverse,kulikov2018importance,tam2019clustered}.
Many of these approaches show marked improvement in diversity over standard beam search across a variety of generative tasks.
However, there has been little attempt to compare and evaluate these strategies against each other on any single task.

In this paper, we survey existing methods for promoting diversity in order to systematically investigate the relationship between diversity and perceived quality of output sequences of conditional language models.
In addition to standard beam search and greedy random sampling, we compare several recently proposed modifications to both methods.
In addition, we propose the use of over-sampling followed by post-decoding clustering to remove similar sequences.

The main contributions of this paper can be summarized as follows:
\begin{itemize}
    \item A detailed comparison of existing diverse decoding strategies on two tasks: open-ended dialog and image captioning, and recommendations for a diverse decoding strategy.
    \item A novel clustering-based algorithm that can be used on the results of any decoding strategy to increase quality and diversity.\footnote{Code can be found at \url{https://github.com/rekriz11/DeDiv}.}
\end{itemize}

\section{Standard Decoding Methods}

Conditional language models, which have wide applications across machine translation, text simplification, conversational agents, and more, generally consist of an encoder, which transforms some input $\textbf{x}$ into a fixed-size latent representation, and a decoder which transforms these representations in order to output a conditional probability of each word in the target sequence given the previous words and the input. 
Let $z_t = f(y_{1},\ldots,y_{t-1}, \mathbf{x})$ represent the output of an encoder-decoder model given input $\mathbf{x}$ and the sequence of tokens predicted so far, $y_1,\ldots,y_{t-1}$, which for notational simplicity we write as $y_{<t}$.
The output $z_t \in \mathds{R}^V$ (where $V$ is the cardinality of the enumerated vocabulary $\mathcal{V}$)

The probability distribution over the next possible token being word $w_i \in \mathcal{V}$ is the softmax:
\begin{align*}
    P(y_t = w_i | y_{<t}, \mathbf{x}) &= \frac{\exp(z_{t,i})}{\sum_{j=1}^{V}{\exp{(z_{t,j})}}}\quad \\
    &\forall i \in \{1,\ldots,V\}
\end{align*}
Most decoding strategies strive to find the most likely overall sequence, i.e. pick a $\mathbf{\hat{y}}$ such that:
\[\Scale[0.85]{
    \mathbf{\hat{y}} = \arg\max_{\mathbf{y}}{ P(\mathbf{y} | \mathbf{x})} = \arg\max_{\mathbf{y}}{\prod_{t=1}^{N} {P(y_t \mid y_{<t}, \mathbf{x})}}
}\]
Unlike Markovian processes, no sub-exponential algorithm exists to find the optimal decoded sequence, and thus we instead use approximations.

\noindent\textbf{Arg-max}\quad The simplest approach to decoding a likely sequence is to greedily select a word at each timestep:
\begin{equation*}
    \hat{y}_t = \arg\max_{y_t}{ P(y_t | y_{<t}, \mathbf{x})}
\end{equation*}
However, because this deterministic approach typically yields repetitive and short output sequences, and does not permit generating multiple samples, it is rarely used in language modelling.

\noindent\textbf{Random Sampling}\quad Another option is to randomly sample from the model's distribution at every timestep. Often, a temperature parameter $T$ is added to control the entropy of the distribution before sampling.

\begin{align*}
    P(y_t = w_i | y_{<t}, \mathbf{x}) &= \frac{\exp(z_{t,i}/T)}{\sum_{j=1}^{V}{\exp{(z_{t,j}/T)}}} \quad \\
    \forall i \in \{1,\ldots,V\}\\
    \hat{y}_t &\sim y_t
\end{align*}
Choosing a temperature greater than one causes outputs to look increasingly more random, while bringing the temperature less than zero causes sequences to increasingly resemble greedy sampling.

Recently, top-$s$ random sampling has been proposed as an alternative to using temperature. Sampling is restricted to the $s$ most likely tokens at each step \cite{fan2018hierarchical,radford2019language}.
We find that top-$s$ random sampling's hard-restriction on generating low probability words is more effective at controlling the stochasticity of sampled sequences than sampling with temperature.

\begin{table*}
\centering
    \small
    \begin{tabular}{|L{2.6cm} | L{4.5cm} || L{2.7cm} | L{4.3cm} |} \hline
    Method & Description & Method & Description \\ \hline
    Random Sampling & Standard decoding mechanism, greedily samples a token from the distribution at each time step. &
    Random Sampling with Temperature & Before sampling, modify entropy of predicted distribution. \\ \hline
    \makecell[l]{Top-$s$ Random\\Sampling\\\citep{fan2018hierarchical}}& Restrict sampling to the $s$-most likely words in the distribution. (story generation) &
    Beam Search & \makecell[l]{Standard decoding mechanism,\\keeps the top $b$ partial hypotheses\\at every time step.\\(machine translation)}\\ \hline
    NPAD Beam Search \citep{cho2016noisy}& Add random noise to the hidden state of the decoder at each time step. (machine translation) &
    \makecell[l]{Top-$g$ Capping\\Beam Search\\ \citep{li2016mutual}}& Only consider the top $c$ hypotheses from each parent hypothesis at each time step. (machine translation, dialog)\\ \hline 
    Hamming Diversity Beam Search \citep{vijayakumar2016diverse}& \makecell[l]{Penalize new hypotheses that have\\many of the same tokens as\\existing partial hypotheses.\\(image captioning)} &
    \makecell[l]{Iterative Beam Search\\ \citep{kulikov2018importance}}& Run beam search several times, preventing later iterations from generating intermediate states already explored. (dialog)\\ \hline
    \makecell[l]{Clustered Beam\\Search\\\citep{tam2019clustered}}& Initially consider more hypotheses at each time step, and then cluster similar hypotheses together. (dialog) & Post-Decoding Clustering (Ours) & Sample a large number of candidates, and then cluster similar outputs together. \\ \hline
    
    \end{tabular}

    \caption{
    Brief high-level descriptions of each decoding method we consider in this paper. In parentheses we give the applications on which the technique was originally applied.
 }
    \label{tab:methods}
\end{table*}

\noindent\textbf{Beam Search}\quad Beam search approximates finding the most likely sequence by performing breadth-first search over a restricted search space.
At every step of decoding, the method keeps track of $b$ partial hypotheses.
The next set of partial hypotheses are chosen by expanding every path from the existing set of $b$ hypotheses, and then choosing the $b$ with the highest scores.
Most commonly, the log-likelihood of the partial sequence is used as the scoring function.
We use this as our baseline.\footnote{We present the beam search algorithm in the appendix.}

Since beam search only explores a limited portion of the overall search space, it tends to yield multiple variants of the same high-likelihood sequence, sequences that often only differ in punctuation and minor morphological changes \cite{li2016mutual}.  
Therefore, standard beam search is not ideal for producing diverse outputs.

\section{Extensions to Beam Search}

In this section, we will discuss a variety of methods that have been developed recently to eliminate redundancy during decoding and generate a wider range of candidate outputs.

\noindent\textbf{Noisy Parallel Approximate Decoding}\quad
Introduced by \citet{cho2016noisy}, NPAD is a technique than can be applied to any decoding setting.
The main idea is that diversity can be achieved more naturally by taking advantage of the continuous manifold on which neural nets embed language.
Instead of encouraging diversity by manipulating the probabilities outputted from the model, diverse outputs are instead produced by adding small amounts of noise to the hidden state of the decoder at each step.
The noise is randomly sampled from a normal distribution. The variance is gradually annealed from a starting $\sigma_0$ to 0 as decoding progresses (that is $\sigma_t = \frac{\sigma_0}{t}$) under the reasoning that uncertainty is greatest at the beginning of decoding.
NPAD can be used in conjunction with any decoding strategy; following the best results from the original paper, we show results using NPAD with beam search.

Extensions to NPAD have sought to learn the direction in which to manipulate the hidden states using an arbitrary decoding objective \cite{gu2017trainable}.
Since such objectives can be highly domain-specific, we do not evaluate this method.

\noindent\textbf{Top-$g$ Capping}\quad
In beam search, it is often the case that one hypothesis $h$ is assigned a much higher probability than all other hypotheses, causing all hypotheses in the next step to have $h$ as their parent. Following \newcite{li2016mutual} and \newcite{li2016simple}, we add an additional constraint to standard beam search to encourage the model to choose options from diverse candidates.
At each step $t$, current hypotheses are grouped according to the parental hypothesis they come from.
After grouping candidates, only the top $g$ from each grouping are considered. The resulting $b \times g$ candidates are ranked, and the top $b$ are selected as hypotheses for the next beam step.

\noindent\textbf{Hamming Diversity Reward}\quad
\newcite{vijayakumar2016diverse} proposes adding an additional diversity-promoting term, $\theta$, to the log-likelihood before reranking.
This term measures how different a candidate hypothesis $c^{(i)}_{\leq t}$ is from the partial hypotheses selected in the previous step. Let $\mathcal{H}_{t-1} = \{c^{(1)}_{\leq t-1}$, \ldots $c^{(b)}_{\leq t-1}\}$ be these partial hypotheses.
Then the beam search scoring function for the $i$th candidate at timestep $t$ becomes:
\begin{align*}
    \text{score}(c^{(i)}_{\leq t}) = \sum_{j=1}^t \big(\log P(c^{(i)}_j | c^{(i)}_{<j}, \textbf{x})\big) \\+ \lambda\theta(c^{(i)}_{\leq t}, \mathcal{H}_{t-1})
\end{align*}
where $\lambda$ is a tunable hyperparameter. \citet{vijayakumar2016diverse} try a variety of definitions for $\theta$, including embedding diversity and $n$-gram diversity, but they find that Hamming distance, the number of tokens in the candidate sequence which exist in the previously selected partial hypotheses, is most effective. We take the negative of the Hamming distance as $\theta$.

\noindent\textbf{Iterative Beam Search}\quad
In an attempt to improve the size of the search space explored without sacrificing runtime, \newcite{kulikov2018importance} propose an iterative beam search method.
Beam search is run many times, where the states explored by subsequent beam searches are restricted based on the intermediate states explored by previous iterations.
Formally, we can define the set of all partial hypotheses for beam search instance $i$ at time step $t$ as $\mathcal{H}_t^{(i)}$. From here, the search space explored by beam search instance $i$ can be expressed as $S_i = \cup_{t=1}^T \mathcal{H}_t^{(i)}$.
The $i$th beam search is prevented from generating any partial hypothesis that has previously been generated, that is, any hypothesis found in $S_{<i} = \cup_{i^{\prime}=0}^{i-1}S_{i^{\prime}}$.

The authors also attempt a soft inclusion criterion, where any states within $\epsilon$ Hamming distance from a previously explored state are also excluded. During the experimentation of \newcite{kulikov2018importance}, however, the soft-inclusion was found to not be beneficial; thus, we only restrict exact matches of previous states in our implementation.
In practice, this means after the first beam search instance runs as normal, the first step of the second beam search instance will contain the $b$+1 to 2$b$-most likely starting tokens; this pattern holds for the third beam search instance, and so on.

\noindent\textbf{Clustered Beam Search}\quad
Most recently, \newcite{tam2019clustered} proposed a clustering-based beam search method to help condense and remove meaningless responses from chatbots.
Specifically, at each decoding step $t$, this method initially considers the top $2*b$ candidates. From there, each candidate sequence is embedded\footnote{We follow \newcite{tam2019clustered} and used averaged GloVe word embeddings \cite{pennington2014glove}.}, and the embeddings are clustered into $c$ clusters using $K$-means. Finally, we take the top $\frac{b}{c}$ candidates from each cluster. Note that in the case any clusters have size less than $\frac{b}{c}$, we then include the highest-ranked candidates not found after clustering.

\section{Clustering Post-Decoding (PDC)}
\label{section:postcluster}

In the previous section, we discuss several diversity-promoting methods that can be applied during the decoding process.
However, it is also possible to encourage additional diversity post-hoc.
On the task of sentence simplification, after decoding using a large-scale diversity-promoting beam search (beam size 100), \newcite{kriz2019complexity} then clustered similar sentences together to further increase the variety of simplifications from which to choose.
Document embeddings generated via Paragraph Vector \cite{Le2014distributed} were used as the sentence embeddings with which to perform $K$-means. 

In this work, we extend this post-decoding clustering idea in three key ways.
First, we make use of sentence-level embeddings which leverage the pre-trained language representations from the Bidirectional Encoder Representations from Transformers (BERT) \cite{devlin2018bert}.\footnote{BERT sentence-level embeddings were obtained using https://github.com/hanxiao/bert-as-service.}
Second, after clustering, \newcite{kriz2019complexity} took the sentence closest to the centroid of each cluster as the representative candidate; we instead choose the highest-ranked candidate (according to log-likelihood) from each cluster to ensure the best candidates are still selected.
Finally, after performing standard $K$-means clustering, we found that it was often the case that some clusters contained large numbers of good candidates, while others contained very few candidates that are also either ungrammatical or otherwise inferior.
Thus, in our implementation, we remove clusters containing two or fewer sentences, and then sample a second candidate from each of the remaining clusters, prioritizing selecting candidates from larger clusters first.

\section{Experimental Setup}
We evaluate the decoding strategies described in the previous sections under the following settings.
For each of the published beam search algorithms, we choose the hyperparameters that were found to be best in the original publications.\\

\small
\begin{tabular}{l|l}
    \multirow{2}{*}{RS} & Random sampling with temp = 0.5,\\
    & 0.7, 1.0, or 1.0 with top-10 capping.\\
    Standard BS & Standard beam search\\
    Top5Cap BS & Top-$g$ capping with $g=3$\\
    Iter5 BS & Iterative beam search with 5 iterations\\
    HamDiv0.8 BS & Hamming Diversity with $\lambda=0.8$\\
    Cluster5 BS & Clustered beam search with 5 clusters\\
    NPAD0.3 BS & Noisy Decoding with $\sigma_0=0.3$ \\
\end{tabular}
\normalsize
\\

For random sampling, we sample 10 outputs, and with beam-search based methods, we use a beam size of 10 to generate 10 outputs.
In addition, we show results from oversampling then filtering.
We use a beam size of 100 or generate 100 samples through random sampling, and then we select 10 from the 100, either through post-decoding clustering (PDC) or by taking the 10 candidates with highest likelihood. 

We examine these decoding strategies on two tasks: open ended dialog and image captioning.
For each task, we evaluate both the quality and diversity of the 10 outputs from each strategy.

\subsection{Open-ended Dialog Task}
In the dialog domain, we use an LSTM-based sequence-to-sequence (Seq2Seq) model implemented in the OpenNMT framework \cite{opennmt}.
We match the model architecture and training data of \citet{baheti2018generating}.
The Seq2Seq model has four layers each in the encoder and decoder, with hidden size 1000, and was trained on a cleaned version of OpenSubtitles \cite{tiedemann2009news} to predict the next utterance given the previous one.

Evaluation is performed on 100 prompts from the Cornell Movie Dialog Corpus~\cite{danescu2011chameleons}.
These prompts are a subset of the 1000 prompts used in \citet{baheti2018generating}, which were filtered using item response theory for discriminative power.

We report perplexity (PpL), averaged over \textit{all} the top 10 outputs for each example.\footnote{This differs from existing work which computes perplexity over only the top output for each example. For our task we are interested in the quality of all of the generated responses.} Since the quality of open-ended dialog is notoriously difficult to evaluate automatically, we ran a human evaluation task on Amazon Mechanical Turk where annotators were shown a prompt and 5 potential responses generated by any of our decoding methods.
Evaluators were asked to provide binary ratings on fluency, adequacy, and interestingness for each response. Overall, we collected 3 human judgments for each of the top ten responses for each of our decoding methods; in other words, we collected 3,000 judgments per method.\footnote{The full instructions shown on AMT are in the appendix.}

\begin{table*}[t]
    \centering
    \small
    \setlength\tabcolsep{4pt} 
    \begin{tabular}{|lr||ccc||ccccc|} \hline
    
    \multicolumn{2}{|c||}{\textbf{Method}} & \textbf{Fluency} & \textbf{Adequacy} & \textbf{Interestingness} & \textbf{Ppl} & \textbf{Dist-1} & \textbf{Dist-2} & \textbf{Ent-2} & \textbf{Ent-4} \\ \hline\hline
    Reference & & 0.795 & 0.732 & 0.636 & -- & -- & -- & -- & -- \\ \hline\hline
    RS 0.7 &(sample 10) & \textbf{0.758} & 0.399 & \textbf{0.388} & 35.98 & 0.63 & 0.80 & 4.08 & 3.84 \\
    RS 1.0 &(sample10) & 0.550 & 0.303 & \td{0.386} & 67.99 & \textbf{0.74} & \textbf{0.87} & \textbf{4.35} & \textbf{4.08} \\
    RS 1.0,top10 &(sample 10) & \td{0.745} & \textbf{0.418} & \td{0.387} & \textbf{10.33} & 0.60 & 0.80 & 4.12 & 3.91 \\ \hline\hline
    Standard BS &(10 beams) & \textbf{0.950} & \textbf{0.621} & 0.336 & \textbf{4.01} & 0.37 & 0.45 & 3.16 & 3.01 \\
    Top3Cap BS &(10 beams)& \td{0.942} & 0.603 & 0.346 & 4.03 & 0.37 & 0.46 & 3.17 & 3.03 \\
    Iter5 BS &(10 beams)& 0.903 & 0.520 & 0.335 & 5.42 & \textbf{0.62} & \textbf{0.74} & \textbf{3.68} & \textbf{3.25} \\
    HamDiv0.8 BS &(10 beams)& 0.923 & 0.599 & \td{0.366} & 4.56 & 0.33 & 0.37 & 3.08 & 3.00 \\
    Cluster5 BS &(10 beams)& 0.936 & 0.582 & \textbf{0.381} & 4.23 & 0.39 & 0.46 & 3.24 & 3.06 \\
    NPAD0.3 BS &(10 beams) & \td{0.942} & \td{0.604} & 0.335 & 4.05 & 0.36 & 0.44 & 3.13 & 2.99 \\ \hline\hline
    RS 1.0,top10 &(sample 100, rank) & \textbf{0.922} & \textbf{0.548} & 0.347 & \textbf{5.10} & 0.52 & 0.68 & 3.54 & 3.18 \\
    RS 1.0,top10 &(sample 100, PDC) & 0.852 & 0.494 & \textbf{0.372} & 6.96 & \textbf{0.63} & \textbf{0.76} & \textbf{3.74} & \textbf{3.27} \\ \hline\hline
    Standard BS &(100 beams, rank) & \textbf{0.964} & \textbf{0.611} & \td{0.332} & \textbf{4.01} & 0.44 & 0.61 & 3.33 & 3.05 \\
    Standard BS &(100 beams, PDC) & 0.944 & 0.599 & \textbf{0.346} & 4.42 & \textbf{0.57} & \textbf{0.70} & \textbf{3.59} & \textbf{3.21} \\ \hline
    \end{tabular}

    \caption{
    Results on 100 dialog prompts.
    The first row shows the mean human ratings of the single reference response available for each prompt.
    The next three rows show results for random sampling, with 10 samples drawn per prompt. The next six rows are variants of beam search using beam size 10.
    The last four rows use random sampling or standard beam search to generate 100 outputs, then filter down to 10 outputs either through ranking by log-likelihood or by performing post-decoding clustering (PDC).
    In each section, the highest value is bolded, and statistical ties are marked \textdagger.}
    \label{tab:results_no_cluster}
\end{table*}

\begin{table*}
    \small
    \centering
\begin{tabular}{|lr||ccc||cccc|}
\hline
 & & \multicolumn{3}{c||}{\textbf{SPICE}} & & & & \\
\multicolumn{2}{|c||}{\textbf{Method}} & \textbf{Mean} & \textbf{@1} & \textbf{@10} & \textbf{Dist-1} & \textbf{Dist-2} & \textbf{Ent-2} & \textbf{Ent-4}  \\
\hline\hline
RS 0.7 &(sample10)               & \textbf{0.170} & \textbf{0.192} & \textbf{0.278} & 0.31 & 0.52 & 3.67 & 4.00 \\
RS 1.0 &(sample10)               & 0.133 & 0.167 & 0.247 & \textbf{0.44} & \textbf{0.71} & \textbf{4.17} & \textbf{4.26} \\
RS 1.0,top10 &(sample10)         & 0.159 & 0.183 & 0.272 & 0.33 & 0.59 & 3.90 & 4.17 \\
\hline \hline
Standard BS &(10 beams)               & 0.194 & 0.193 & 0.283 & 0.18 & 0.26 & 2.94 & 3.18 \\
Top3Cap BS &(10 beams)                & \textbf{0.195} & \textbf{0.196} & 0.282 & 0.17 & 0.26 & 2.93 & 3.17 \\
HamDiv0.8 BS &(10 beams)              & 0.194 & 0.194 & 0.282 & 0.18 & 0.27 & 2.98 & 3.19 \\
Cluster5 BS &(10 beams)               & 0.191 & 0.194 & \textbf{0.285} & \textbf{0.19} & \textbf{0.28} & \textbf{3.04} & \textbf{3.25} \\
NPAD0.3 BS &(10 beams)                & 0.191 & 0.192 & 0.280 & 0.18 & 0.26 & 2.94 & 3.17 \\
\hline \hline
RS 1.0,top10 &(sample100, rank)  & \textbf{0.182} & \textbf{0.188} & \textbf{0.284} & 0.25 & 0.41 & 3.31 & 3.64 \\
RS 1.0,top10 &(sample100, PDC)   & 0.169 & \textbf{0.188} & 0.282 & \textbf{0.31} & \textbf{0.52} & \textbf{3.62} & \textbf{3.91} \\
\hline \hline
Standard BS &(100 beams, rank)        & \textbf{0.188} & 0.190 & 0.279 & 0.20 & 0.31 & 3.04 & 3.32 \\
Standard BS &(100 beams, PDC)         & 0.186 & \textbf{0.192} & \textbf{0.288} & \textbf{0.24} & \textbf{0.38} & \textbf{3.25} & \textbf{3.57} \\
\hline
\end{tabular}
\caption{Image captioning results for selected random sampling and beam search methods. SPICE@1 measures the SPICE score of the most likely caption. SPICE@10 is the maximum score across the 10 candidates generated by each method. Mean SPICE is the mean score over all 10 candidates. In each section, the best value is bolded.}
\label{tab:image_captioning}
\end{table*}

\subsection{Image Captioning Task}
For image captioning, we use a state-of-the-art model introduced in \citet{anderson2018bottom}.
We take advantage of \citet{Luo2017}'s open-source implementation and released model parameters trained on MSCOCO \cite{lin2014microsoft}.
We evaluate on a test set containing 5000 images.

We report Semantic Propositional Image Caption Evaluation (SPICE) scores, an automatic evaluation metric that has been shown to correlate well with human judgments of quality\citep{Anderson2016SPICE}. 
SPICE measures how well the semantic scene graph induced by the proposed caption matches one induced by the ground truth.
In addition to computing SPICE on the top-scoring caption (SPICE@1), we follow \citet{vijayakumar2016diverse} in reporting Oracle SPICE@10 scores.
This is done to show the upper bound on the potential impact diversity can have.
We also compute the mean SPICE score across all of the candidate captions for an image.
Unlike SPICE@1 and SPICE@10, this metric shows the overall quality of \textit{all}
of the candidate captions, which is useful to know for applications that combine diverse candidate output sequences \cite{krause2017hierarchical}.

\subsection{Evaluating Diversity}
To measure the diversity across the generated candidate sequences for a given input, we report \textbf{Dist-k}, the total number of distinct k-grams divided by the total number of produced tokens in all of the candidate responses for a prompt \citep{li2016diversity}. 
We report Dist-2 and Dist-4 averaged over the prompts in the test set.

A limitation of Dist-$k$ is that all $k$-grams that appear at least once are weighted the same, ignoring the fact that infrequent $k$-grams contribute more to diversity than frequent ones. 
\citet{zhang2018generating} instead propose an entropy metric, \textbf{Ent-k}, defined as:
 \begin{align*}
 \textit{Ent-k} = \frac{-1} {\sum_{w \in S}F(w)} \sum_{w \in S} F(w) \log \frac{F(w)} {\sum_{w' \in S} F(w')}
 \end{align*}
 where $S$ is the set of all $k$-grams that appear in candidate responses for an example, and $F(w)$ denotes the frequency of $w$ in the candidate responses.

\section{Results}
We report results on dialog systems and image captioning in Tables \ref{tab:results_no_cluster} and \ref{tab:image_captioning}, respectively. As expected, random sampling-based approaches yield outputs with greater diversity but worse quality than beam search-based approaches.
Over-sampling then filtering increases the quality of outputs while still ensuring high diversity. 
In the following sections, we discuss the diversity-quality tradeoff, and then delve further into the results for each method group.

\begin{figure*}[ht]
    \centering
    \begin{subfigure}[l]{5cm}
        \includegraphics[width=5cm]{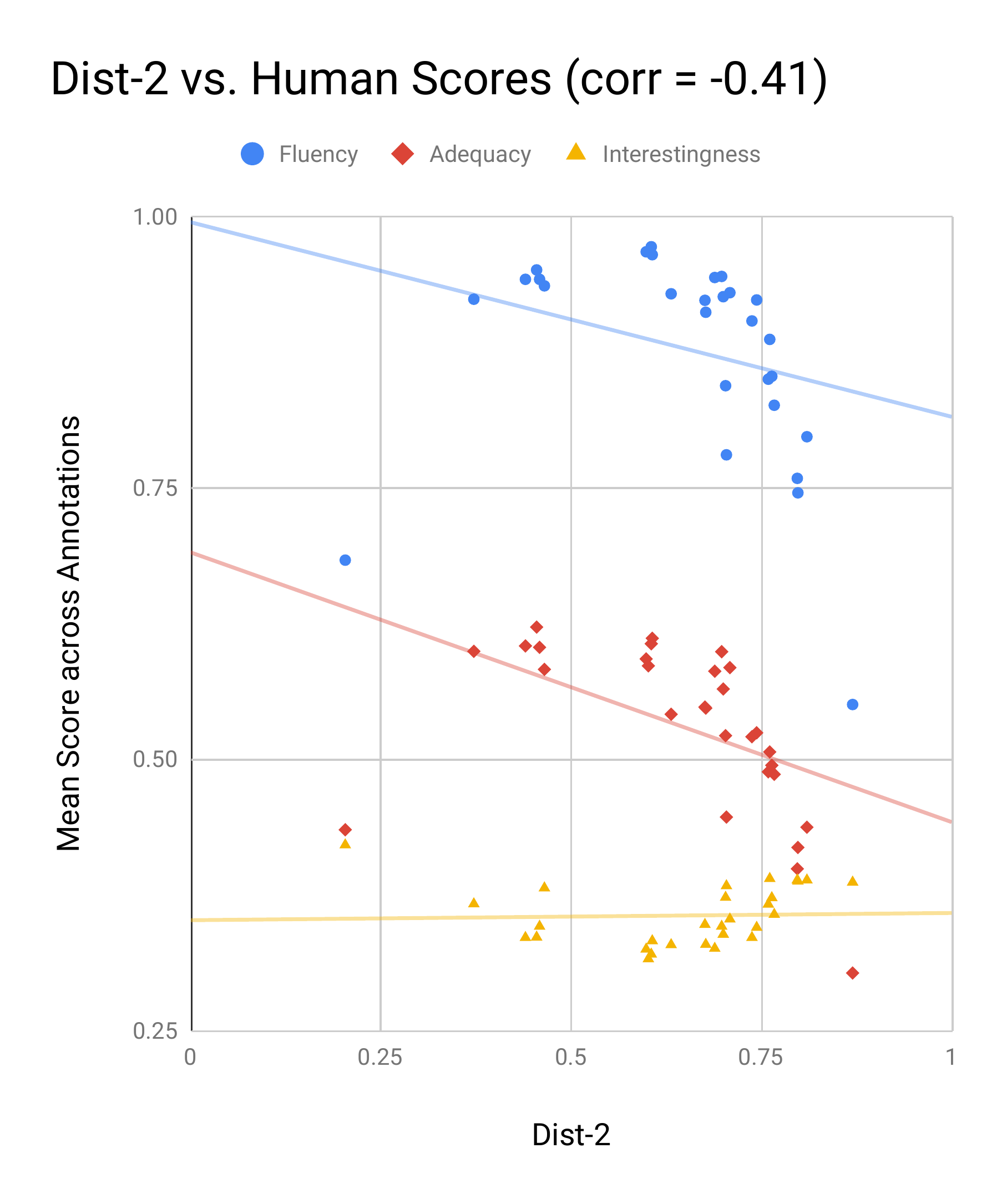}
    \end{subfigure}
    \begin{subfigure}[l]{5cm}
        \includegraphics[width=5cm]{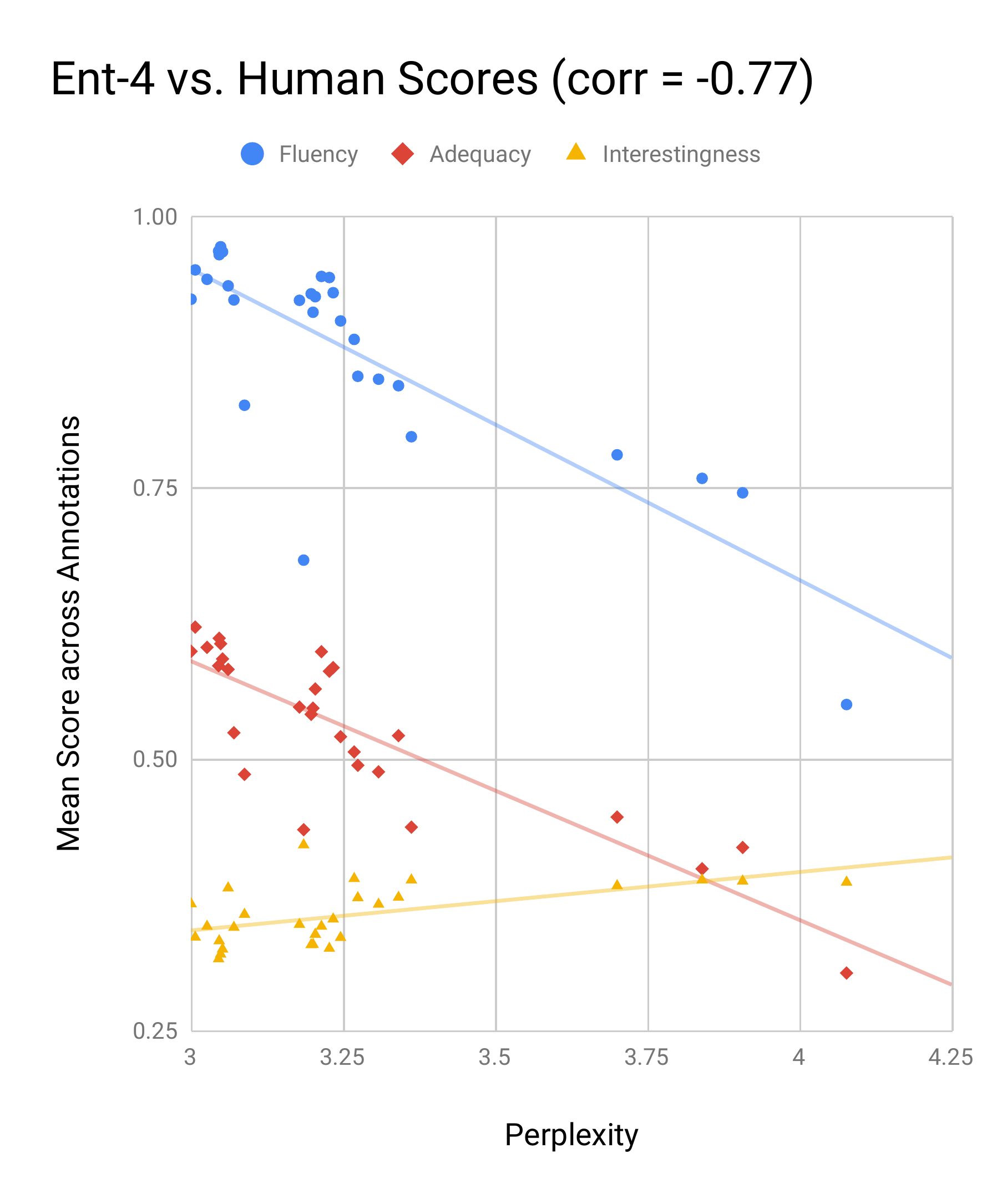}
    \end{subfigure}
    \begin{subfigure}[l]{5cm}
        \includegraphics[width=5cm]{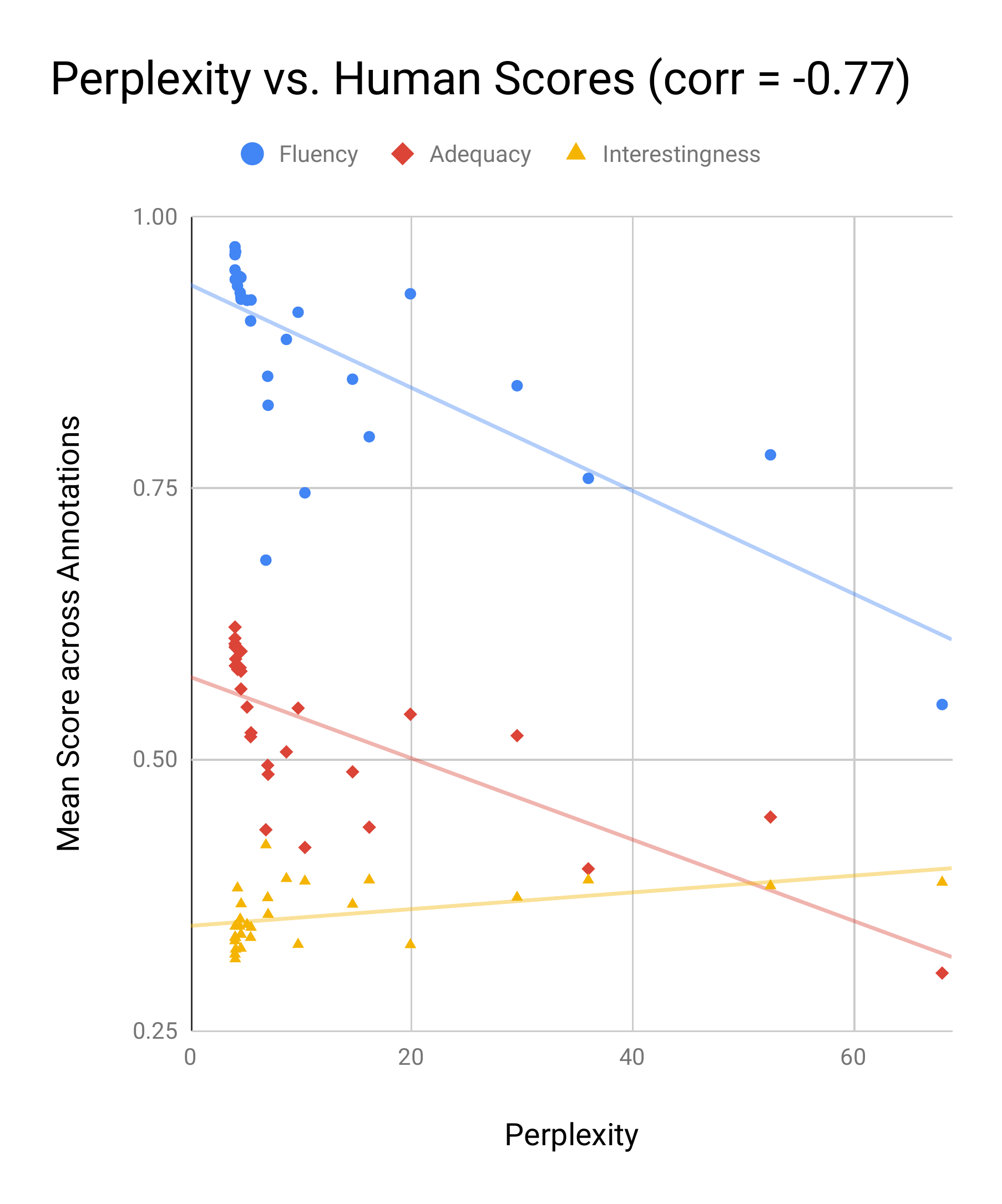}
    \end{subfigure}
    \caption{Each decoding strategy is plotted, showing that human-perceived quality is negatively correlated with diversity. The Pearson Correlation coefficients between each statistic and the average of fluency, coherence, and interestingness are shown in parentheses.}
    \label{fig:correlations}
\end{figure*}

\begin{table*}[t]
    \centering
    \small
    \begin{tabular}{|l|l|l|}
    \hline
    \multicolumn{3}{|l|}{\rule{0pt}{0.35cm} \textbf{Prompt:} Look, nobody knows we did it.}  \\ 
    \hline \hline
    \textbf{RS 0.5} &  \textbf{Standard BS} &  \textbf{NPAD0.3 BS} \\
    \hline
\makecell[l]{
I don't know what you're talking about. \\
What's the matter with you? \\
I don't know what it is. \\
I don't think so. \\
He's got to get out of here. \\
}&
\makecell[l]{
We've got to get out of here. \\
What do you mean? \\
I don't think it's a good idea. \\
I don't know what to say. \\
I don't know what's going on. \\
}&
\makecell[l]{
I don't think it's a good idea. \\
I don't know what to say. \\
I don't know what's going on. \\
I don't know what to do. \\
I don't know what's going on here. \\
} \\
    \hline
    \hline
    \textbf{RS 1.0} & \textbf{Standard BS with PDC} &  \textbf{Cluster5 BS} \\
    \hline
    
\makecell[l]{
I can't find it. \\
They're our ships. \\
It's all right anyone is the right to interfere. \\
We didn't have a plan I engineered a policy. \\
Same time you pick us up at six and get we. \\
}&
\makecell[l]{
I don't know! \\
I don't think so. \\
What do you mean? \\
Why didn't you tell me? \\
That's why we're here. \\
}&
\makecell[l]{
I don't know why. \\
What do you mean? \\
I don't think so. \\
How do you know that? \\
I'll tell you what. \\
}
\\
    \hline
    \hline
    \textbf{RS 1.0,top10} & \textbf{RS 1.0,top10 with PDC} &  \textbf{Top3Cap BS} \\
    \hline
\makecell[l]{
I don't know what else to do. \\
It doesn't have to be that way! \\
We're in the air! \\
I've seen a guy in his place in a it. \\
And I'm not we any more. \\
}&
\makecell[l]{
What do you mean? \\
I don't think so. \\
That's why I'm here. \\
It's all right we. \\
We've been through this before. \\
}&
\makecell[l]{
We've got to get out of here. \\
What do you mean? \\
I don't think it's a good idea. \\
I don't know what to say. \\
I don't know what's going on. \\
} \\
\hline
    \end{tabular}
    \caption{Responses to an example prompt for selected methods. More examples can be seen in the appendix.}
    \label{examples}
\end{table*}

\subsection{The Quality Diversity Tradeoff}

The goal of diverse decoding strategies is to generate high-quality candidate sequences which span as much of the space of valid outputs as possible. 
However, we find there to be a marked trade-off between diversity and quality.
This can be seen in Figure \ref{fig:correlations}, where we plot the human-judged quality score for each dialog experiment against our primary diversity descriptive statistics.
Fluency and adequacy are both strongly negatively correlated with diversity.
While we had expected interestingness to be positively correlated with diversity, the fact that it is not suggests that existing diversity statistics are insufficient for capturing what it means to humans for outcomes to be interesting.

Likewise, in image captioning, the mean SPICE score of the 10 candidate captions (averaged over all examples for each experimental setting) is strongly anti-correlated with diversity, with a Pearson correlation coefficient of -0.83 with the Ent-4 measure and -0.84 with Dist-2.
Clearly it remains an open challenge to generate a diverse set of image captions that are all high-quality.

When researchers choose to use a diverse decoding strategy, they must decide where on the quality-diversity tradeoff they would like to lie; selecting an optimal method depends strongly on one's tolerance for errors.
In machine translation, where mistakes could severely impact coherence, beam search-based methods, which tend to result in better fluency and coherence, but worse diversity might be preferred.  
In more open-ended applications, where novel text is of greater importance, increased diversity could be worth the fluency and coherency hit.
As state-of-the-art models continue to improve, one would hope that the quality cost of encouraging diversity will continue to decrease. 

In the interest of reporting a single overall best method for each task, we computed a sum-of-ranks score for each method.
For dialog, we ranked the methods each by fluency, coherence, interestingness, and Ent-4, and then took a weighted sum of the four ranks, with 50\% of the weight assigned to Ent-4, and 50\% distributed evenly among the human evaluation ranks.
Overall, clustered beam search and standard BS (beam size 100, PDC) have the best scores, followed by clustered beam search (beam size 10).
Similarly, for image captioning, we rank the methods by their mean SPICE score and by Ent-4.
Summing these ranks, random sampling (temp 1.0, top-10 capping, PDC) came in first.
Standard beam search, Hamming Diversity beam search, and Top-$g$ capping beam search (beam size 10) tied for second.

\subsection{Random Sampling-based Methods}
Higher sampling temperatures result in both an increase in diversity in generated responses and a reduction in overall quality.
In the dialog domain, evaluators consistently rate the responses sampled with temperature 1.0 to have worse fluency, coherence, and interestingness when those sampled with temperature 0.5.
In the image captioning domain, lower temperature improves automatic evaluation metrics for quality while reducing diversity.

For dialog, restricting sampling to the top-10 vocabulary words is a more effective strategy than adjusting temperature for ensuring balance between the quality and diversity of outputs.
Top-10 random sampling has the highest fluency, coherence, and interestingness, as well as significantly lower perplexity than other random sampling methods.
However, this trend did not extend to image captioning, where top-10 random sampling results in both worse SPICE scores and lower diversity measures than setting the temperature to 0.7.
This may be because image captioning is a less ambiguous task than open-ended dialog, leading to a better-trained model that puts more probability mass on high-quality vocabulary words, ameliorating the challenge top-$c$ filtering is designed to eliminate: that of a long tail of low probability vocabulary words taking up a large amount of probability mass.

\subsection{Beam Search-based Methods}

For dialog, clustered beam search (Cluster5 BS) performs the best of all beam search methods in terms of human-judged interestingness. It ties for best with NPAD0.3BS on fluency and ties with Standard BS on coherence.
Iterative beam search (Iter5 BS) achieves the greatest diversity, but at the expensive of quality.
It has the lowest human-judged coherence among beam search methods; thus, we do not evaluate this method on image captioning.
For image captioning, Cluster5 BS has the highest diversity among beam search methods, but this difference is quite small.
Cluster5 BS also has the highest SPICE@10 score, indicating it is the best method for generating at least one high quality candidate.
However, Top3Cap BS results in the highest mean SPICE score, suggesting it is best at ensuring all outputs are reasonable quality.
% This indicates that for rereanking tasks 

\subsection{Effect of Over-sampling}

In our experiments, we explore over-sampling 100 outputs, and then either using post-decoding clustering (PDC) or re-ranking by log-likelihood to filter these 100 down to 10 diverse outputs.

In the dialog domain, this over-sampling approach is a definite win.
When over-sampling with random sampling both methods of filtering substantially improve human judgements of fluency and adequacy compared to random sampling only 10 outputs.
However, interestingness scores go down, and while the outputs are still more diverse than beam search-based methods, they are less diverse than random sampling without filtering.
In the beam search methods that use a beam size of 100 then filter down to 10, human-judged quality is on par with beam size 10 results, but diversity is considerably higher.

When comparing the two types of filtering, PDC results in higher interestingness and diversity statistics, while log-likelihood re-ranking improves fluency and adequacy.
This again demonstrates the trade-off between quality and diversity.\footnote{In the appendix, we show results with every method where we generate 10 samples; generate 100 samples followed by selecting the 10 most likely outputs; and generate 100 samples followed by post-decoding clustering to select 10 outputs.}

For image captioning, over-sampling with reranking does not consistently improve quality as it does in the dialog domain.
Mean SPICE score is improved for random sampling but not for beam search. 
SPICE@1 becomes worse for both random sampling and decoding, while SPICE@10 improves for random sampling, and for beam search when PDC is applied.
From these results, we can conclude that over-sampling then ranking does not have a sizeable effect, either negative or positive, on quality.
Moreover, the diversity of the captions generated by random sampling actually decreases when oversampling. The diversity of beam search-generated captions does improve with over-sampling.

While oversampling does generally improve outcomes on the diversity/quality tradeoff, it is more computationally expensive, particularly with beam search.
Running PDC also requires generating sentence embeddings for every output, which adds additional computation time.

\section{Additional Related Work}
In this paper, we have compared a variety of post-training diversity-promoting algorithms. Here, we discuss other related works that instead promote diversity at train-time, as well as alternative quality evaluation methods. We also note that concurrent work has proposed nucleus sampling as an improvement to the sampling strategies discussed in this paper \cite{holtzman2019curious}.

\textbf{Diversity Promotion During Training}\quad
Several works have attempted to encourage diversity during training by replacing the standard log-likelihood loss with a diversity-promoting objective.
\citet{li2016diversity} introduces an objective that maximizes mutual information between the source and target.
\citet{zhang2018generating} uses an adversarial information maximization approach to encourage generated text to be simultaneously informative and diverse.
\citet{xu2018diversity} also uses an adversarial loss; their loss function rewards fluent text and penalizes repetitive text. 
We do not evaluate on these methods as they tend to be task-specific and difficult to implement.
All of the diversity strategies we evaluate share the trait that they are agnostic to model architecture and to the data type of the input, as long as the output of the model is a probability distribution over tokens in a sequence.

\textbf{Automatic Quality Evaluation}\quad
An important part of this work is how to accurately measure not only the effect these methods have on candidate diversity, but also on the overall quality of the candidates.
In choosing to report human scores and perplexity for the dialog domain, and SPICE for image captioning, we omitted some quality measures used in other papers.

For image captioning, BLEU \citep{Papineni2001Bleu}, ROUGE \citep{lin2004rouge}, METEOR \citep{elliott2013image}, and CIDer \citep{Vedantam2015Cider} scores are often reported, but SPICE has been shown to have higher correlation with human judgments \cite{Anderson2016SPICE}. 
In the dialog domain, single-reference BLEU score \citep{Papineni2001Bleu} is sometimes used to measure response quality, but it has been shown to have little correlation with human-judged quality \citep{liu2016not}.
Therefore, most works in dialog systems use human evaluation as the ultimate measure of quality \citep{li2016diversity,sedoc2018chateval}

\section{Conclusion}

In this work, we perform an analysis of post-training decoding strategies that attempt to promote diversity in conditional language models.
We show how over-sampling outputs then filtering down to the desired number is an easy way to increase diversity.
Due to the computational expense of running large beam searches, we recommend using random-sampling to over-sample.
The relative effectiveness of the various decoding strategies differs for the two tasks we considered, which suggests that choice of optimal diverse decoding strategy is both task-specific and dependent on one's tolerance for lower quality outputs.

While we have focused on evaluating each decoding strategy under the specifics reported to be the best in the original, further work is necessary to conclude whether observed differences in quality and diversity may simply be due to each work's chosen hyperparameters.
The ability to effectively generate a diverse set of responses while not degrading quality is extremely important in a variety of generation tasks, and is a crucial component to harnessing the power of state-of-the-art generative models.

\section{Acknowledgements}
We thank our anonymous reviewers for helpful feedback.
We also thank Yun William Yu for assistance with statistical testing and proofreading.
This material is based in part on research sponsored by DARPA under grant number HR0011-15-C-0115 (LORELEI). The U.S. Government is authorized to reproduce and distribute reprints for Governmental purposes. The views and conclusions in this publication are those of the authors and should not be seen as representing official endorsements of DARPA and the U.S. Government.

\bibliography{naaclhlt2019}
\bibliographystyle{acl_natbib}

\end{document}